\documentclass[10pt,twocolumn,letterpaper]{article}

\usepackage{iccv}
\usepackage{times}
\usepackage{epsfig}
\usepackage{graphicx}
\usepackage{amsmath}
\usepackage{amssymb}
\graphicspath{{Figures/}}
\usepackage[inline]{enumitem}
\usepackage{multirow}
\usepackage{fixltx2e}
\usepackage{floatrow}
\usepackage{subfig}
\usepackage{soul}
\usepackage{algorithmic}
\usepackage[ruled,vlined]{algorithm2e}
\usepackage{makecell}
\usepackage{authblk}

\usepackage[pagebackref=true,breaklinks=true,letterpaper=true,colorlinks,bookmarks=false]{hyperref}
\iccvfinalcopy 

\def\iccvPaperID{2117} 

\ificcvfinal\fi
\newlist{multiitem}{itemize}{1}
\setlist[multiitem]{
	label=\textbullet,
	before=\begin{multicols}{2},
		after=\end{multicols}
}

\def\etal{\emph{et al.}}

\makeatletter
\newcommand{\inlineitem}[1][]{%
	\ifnum\enit@type=\tw@
	{\descriptionlabel{#1}}
	\hspace{\labelsep}%
	\else
	\ifnum\enit@type=\z@
	\refstepcounter{\@listctr}\fi
	\quad\@itemlabel\hspace{\labelsep}%
	\fi}
\begin{document}

\title{Network Architecture Search for Face Enhancement}

\author[1]{Rajeev~Yasarla\thanks{Work performed during internship at Microsoft}}
\author[2]{Hamid~Reza~Vaezi~Joze}
\author[1]{Vishal~M.~Patel}
\affil[1]{Department of Electrical and Computer Engineering, Johns Hopkins University}
\affil[2]{Microsoft}
\affil[ ]{\{ryasarl1,vpatel36\}@jhu.edu, hava@microsoft.com} 
\maketitle
\begin{abstract}

Various factors such as ambient lighting conditions, noise, motion blur, etc. affect the quality of captured face images.  Poor quality face images often reduce the performance of face analysis and recognition systems.  Hence, it is important to enhance the quality of face images collected in such conditions.  We present a multi-task face restoration network, called Network Architecture Search for Face Enhancement (NASFE), which can enhance poor quality face images containing a single degradation (i.e. noise or blur) or multiple degradations (noise+blur+low-light).  During training NASFE uses clean face images of a person present in the degraded image to extract the identity information in terms of features for restoring the image.  Furthermore, the network is guided by an identity-loss so that the identity information is maintained in the restored image.	Additionally,  we propose a network architecture search-based fusion network in NASFE which fuses the task-specific features that are extracted using the task-specific encoders. We introduce FFT-op and deveiling operators in the fusion network to efficiently fuse the task-specific features. Comprehensive experiments on synthetic and real images  demonstrate that the proposed method outperforms many recent state-of-the-art face restoration and enhancement methods in terms of quantitative and visual performance.

\end{abstract}

\section{Introduction}
\begin{figure}[ht!]
	\centering
	\includegraphics[width=0.24\textwidth,height = 0.30\textwidth]{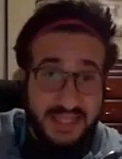}
	\includegraphics[width=0.24\textwidth,height = 0.30\textwidth]{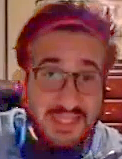}
	\includegraphics[width=0.24\textwidth,height = 0.30\textwidth]{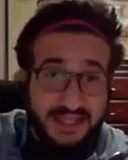}
	\includegraphics[width=0.24\textwidth,height = 0.30\textwidth]{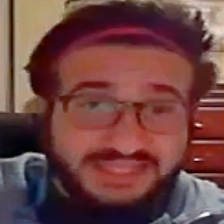}
	\\ \vskip 1pt
	\begin{flushleft}
		\vskip-15pt
		{\footnotesize \hskip3pt Degraded image  \hskip12pt Super-FAN~\cite{bulat2018super} \hskip12pt Shen~\textit{et al.}~\cite{ZiyiDeep} \hskip10pt UMSN~\cite{yasarla2020deblurring} }\\ 
	\end{flushleft}
	\vskip-7pt
	\centering
	\includegraphics[width=0.24\textwidth,height = 0.30\textwidth]{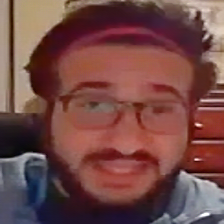}
	\includegraphics[width=0.24\textwidth,height = 0.30\textwidth]{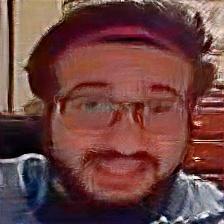}
	\includegraphics[width=0.24\textwidth,height = 0.30\textwidth]{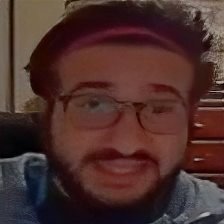}
	\includegraphics[width=0.24\textwidth,height = 0.30\textwidth]{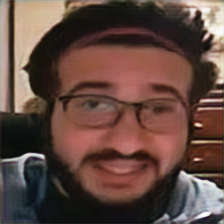}\\
	\begin{flushleft}
		\vskip-15pt
		{\footnotesize  DeblurGANv2\cite{kupyn2019deblurgan}  \hskip5pt DFDNet~\cite{Li_2020_ECCV} \hskip8pt HiFaceGAN~\cite{Yang2020HiFaceGANFR} \hskip6pt NASFE (ours)}\\ 
	\end{flushleft}
	\vskip-20pt
	\caption{Sample results on the face with multiple degradations like blur, noise and low-light conditions.  Restoration methods such as  \cite{PanECCV14,ZiyiDeep,bulat2018super,yasarla2020deblurring,Li_2020_ECCV,Yang2020HiFaceGANFR} fail to reconstruct a high quality clean face image. 
		In constrast, the proposed NASFE network  produces a high quality face image. }
	\label{Fig:real_img1}
\end{figure}

	In the era of COVID-19, the use of video communication tools such as Zoom, Skype, Webex, MS Teams, Google Meet etc. has increased drastically.   In many cases, images/videos captured by these video conferencing tools are of poor quality due to low-light ambient conditions, noise, motion artifacts etc. Figure~\ref{Fig:real_img1} shows an example of such an image taken during a video  conference.  Hence, it is important to enhance the quality of face images collected in such conditions.  Furthermore, restoration of degraded face images is an important problem in many applications such as human computer interaction (HCI), biometrics, authentication  and surveillance.

	Existing face image restoration and enhancement methods are designed to address only a single type of degradation such as blur, noise or low-light.  However, in practice face images might have been collected in the presence of multiple degradations (i.e. noise + blur+ low-light).  Hence, it is important to enhance the quality of face images collected in such conditions.  In this paper, we address the problem of restoring a single face image degraded by multiple degradations (noise+blur+low-light).  To the best of our knowledge, we are first ones to address such a multi-task image restoration problem where a single network is able to remove the effects of low-light conditions, noise, and blur simultaneously.


Face images are structured and informative when compared to natural images.  Methods such as  \cite{PanECCV14,ZiyiDeep,bulat2018super,yasarla2020deblurring}  extract structured information from faces in the form of semantic maps or exemplar masks in order to super-resolve or remove blur from the degraded face images.  This way of extracting facial semantic information is extremely difficult when there are multiple degradations present in the image  and as a result may lead to poor restoration performance. Recently, HiFaceGAN~\cite{Yang2020HiFaceGANFR} addressed face restoration using a multi-stage semantic generation framework.  The performance of this method relies heavily on the features extracted from the degraded faces for capturing  semantic information.  However, it is difficult to learn  semantic structure from the noisy features corresponding to multiple degradations.  Recently, DFDNet~\cite{Li_2020_ECCV} proposed a dictionary-based method to produce a high-resolution face image from a low-resolution input.  Note that super resolution is a relatively easy task  compared to restoring an image from multiple degradations since semantic information cannot be easily extracted from the images with multiple degradations.  As shown in Figure~\ref{Fig:real_img1} even though state-of-the-art face image restoration methods such as \cite{PanECCV14,ZiyiDeep,bulat2018super,yasarla2020deblurring,Li_2020_ECCV,Yang2020HiFaceGANFR}  are retrained on multiple degradations, they fail to reconstruct a high-quality clean  face image. 

Recently, Li \etal~\cite{Li_2020_CVPR} proposed a blind face restoration method by utilizing multi-exemplar images and adaptive fusion of features from guidance and degraded images.  An important point to note here is that computation of the guidance image also relies on extracting the structured information (i.e. landmarks) from degraded face image and clean face image.  Since structured information extracted from a degraded  image with multiple degradations is not reliable, this way of computing a guidance image will not be helpful in addressing the proposed problem of multi-task face image restoration. To address this problem, we propose a way of using clean face images corresponding to same person in the degraded image  to restore the image. These clean images, which maybe taken in different scenarios at different times,  help us extract the identity information using VGGFace features \cite{Parkhi15}.  Since these images may have different styles due to contrast or illumination, we apply Adaptive-Instance normalization (AdaIN)~\cite{huang2017arbitrary} on the extracted VGGFace features to transform the style to that of the degraded face image.  We also use these extracted features and propose a novel identity-loss, $\mathcal{L}_{iden}$, for training the network.

\begin{figure*}[t!]
	\begin{center}
		\includegraphics[width=0.85\linewidth, height=8.5cm]{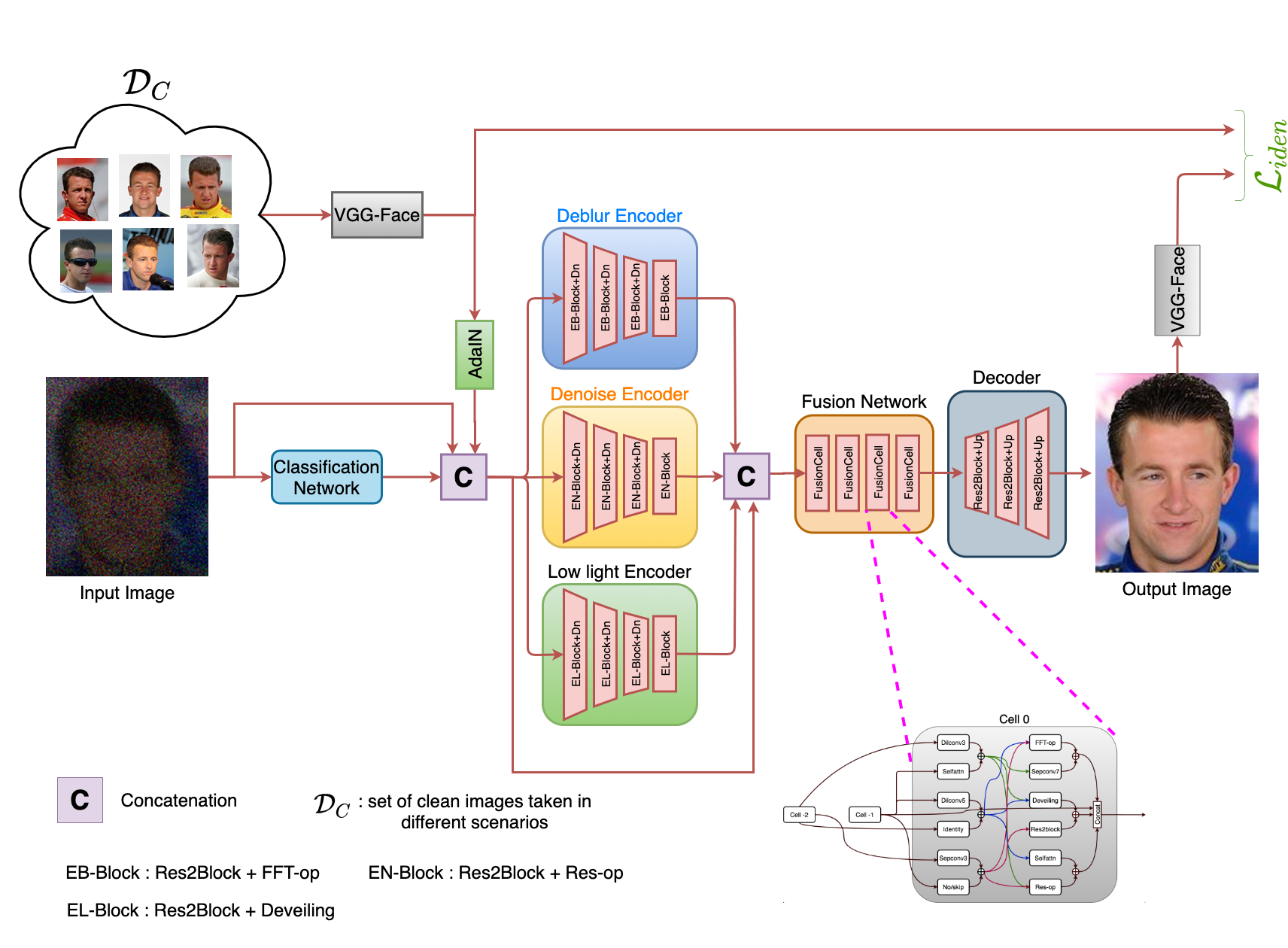}
	\end{center}
	\vskip -20pt 
	\caption{An overview of the proposed NASFE face image restoration network.}
	\label{fig:overview}
\end{figure*}

The proposed multi-task face restoration problem can be considered as an many-to-one feature mapping problem, i.e extracting task specific features (i.e. noisy, blur, and low-light enhancement features) and fusing them to get features corresponding to the clean image.  The fused features can then be used by a decoder to restore the face image. One can clearly see the importance of fusion in this framework.  Rather than  naively using Res2Blocks~\cite{gao2019res2net} or convolutional blocks in the fusion network, one can learn an  architecture for fusion which may lead to better restoration. To this end, we propose a  neural architecture search-based approach \cite{liu2017hierarchical,liu2018darts} for learning  the fusion network architecture.  Additionally, we introduce FFT-op and deveiling operators in order to process the task specific features efficiently where these operators address image formation formulation of these multiple degradations.  FFT-op is motivated by Weiner deconvolution and help in learning the  weights to efficiently fuse the task specific features and remove the effect of blur from the features.  Deveiling operator is introduced to learn the weights in order to enhance low-light conditions efficiently.  Furthermore, we use a classification network to classify the input degraded image into different classes that give  information about the degradations that are present in the input face image. This class specific information  is used as a prior information in the fusion network for fusing the task specific features.  Fig.~\ref{Fig:real_img1} shows sample results from the proposed Network Architecture Search for Face Enhancement (NASFE) method, where one can see that NASFE is able to provide better restoration results as compared to the state-of-the-art face restoration methods.

To summarize, this paper makes the following contributions:
\begin{itemize}[noitemsep]
	\itemsep0em
	\item We propose a way of extracting the identity information form different clean images of a person present in the degraded image to restore the face image.  

	\item We propose a novel loss, called identity-loss ($\mathcal{L}_{iden}$), which uses the aforementioned identity information to train the NASFE network.
	\item We propose a neural architecture search-based method for designing the fusion network.
\end{itemize}

\section{Related work}


\noindent {\bf{Denoising.}}  Earlier methods such as \cite{dabov2007image,zoran2011learning,dong2012nonlocally,gu2014weighted} make use of image priors to perform denoising.  They require  the knowledge about the amount of noise present in the noisy image to obtain denoised images. In contrast to these methods, blind image denoising methods like~\cite{zhu2016noise,lebrun2015multiscale,liu2007automatic}, model the noise using techniques like  Non-local Bayes, and Low-rank mixture of Gaussians. CSF~\cite{schmidt2014shrinkage} and TDRN~\cite{chen2016trainable} proposed optimization algorithms in addressing stage-wise inference methods. The advent of convolutional networks in addressing image restoration problem allowed, DnCNN~\cite{zhang2017beyond}, FFDNet~\cite{zhang2017multistyle}, RED30~\cite{mao2016image}, and BM3D-Net~\cite{yang2017bm3d} to achieve impressive performance in denoising. Noise2Noise~\cite{lehtinen2018noise2noise} method proposed a method that doesn't require paired noise-clean images to train the network.  They rely on statistical reasoning and train the network using pairs of noisy images. Authors of CBDNet~\cite{guo2019toward} proposed an elegant way to model realistic noisy image, and use them to train the network that uses asymmetric learning to suppress under estimation of noise level.

\noindent {\bf{Deblurring.}}
Classical image deblurring methods follow  estimation of blur kernel given a blurry image, and then use a deconvolution technique to obtain deblurred image. \cite{Xu2013,Fergus06removingcamera,Krishnan2011,Ren2016,Schuler2013,Pan_2016_CVPR,patchdeblur_iccp2013,ManifoldDeblur,ShearDeconv} methods have been proposed to compute different priors like sparsity, $L_0$ gradient prior, patch prior, manifold prior, and low-rank prior, in order to compute blur kernel from the given blurry image. In recent years, neural network-based methods have also been proposed for deblurring~\cite{ayan2016,ren2019face,lu2019unsupervised,ZiyiDeep,yasarla2020learning}, super-resolution~\cite{yu2018face,chen2018gated,bulat2018super,Li_2020_CVPR,xu2017learning,vaezi2020imagepairs}. Deblurring methods are classified into blind and non-blind  deblurring methods based on the usage of blur kernel information while restoring the  image. Recent non-blind image deblurring methods like \cite{Xu2014,Kruse2017,Schmidt2013,Schuler2013,Schmidt2014,vasu2018non} assume and use some knowledge about the blur kernel. Several face deblurring methods extract structural information in the form of facial fucidial or key points, face exemplar mask or semantic masks. Pan \etal \cite{PanECCV14} extract exemplar face images and  use them as a global prior to estimate the blur kernel. Recently, \cite{ZiyiDeep,shen2020exploiting} proposed to use the semantic maps of a face to deblur the image. HiFaceGAN~\cite{Yang2020HiFaceGANFR} addressed face restoration using a multi-stage semantic generation framework. DFDNet~\cite{Li_2020_ECCV} proposed a dictionary-based method to produce high-resolution face image.

\noindent {\bf{Low-light enhancement.}} Various methods have been proposed in the literature for addressing  low-light enhancement problem.  Methods such as  histogram equalization~\cite{pizer1987adaptive}, matching region
templates~\cite{hwang2012context}, contrast statistics~\cite{rivera2012content}, bilateral learning~\cite{gharbi2017deep}, intermediate HDR supervision~\cite{yang2018image}, reinforcement learning~\cite{park2018distort,6198348}, and adversarial~\cite{Chen_2018_CVPR,ignatov2017dslr} have been proposed in the literature.  \cite{Dono2005,PatchBasedHDR} view the inverse of low-light image as a hazy image in order to estimate a low-light enhanced image. Wei \etal~\cite{wei2020physics} proposed a realistic noise formation model based on the characteristics of CMOS photosensors, to synthesize realistic low-light images. Guo \etal~\cite{guo2020zero} proposed a zero reference-based method that uses pixel-wise high-order curves for dynamic range adjustment to enhance dark images.

\noindent {\bf{Network Architecture Search.}}
Neural Architecture Search methods focuses on automatically designing or constructing neural network architectures that are efficient in addressing a specific problem to achieve the optimal performance. Several architecture search methods have been proposed that use reinforcement learning~\cite{zoph2016neural,cai2017efficient,zhong2018practical} and evolutionary algorithms~\cite{real2017large,xie2017genetic,liu2017hierarchical,miikkulainen2019evolving}.  Recent architecture search methods like  \cite{pham2018efficient,zoph2018learning,real2019regularized}  focus on searching the repeatable cell structure, while fixing the
network level structure fixed. These methods are more efficient and computationally less expensive than the earlier methods. PNAS~\cite{liu2018progressive} proposed a progressive search strategy that notably reduces the computational cost. Motivated by~\cite{liu2019auto,liu2018darts}, we propose an efficient way to design our fusion network that fuses the task specific features in addressing the removal of multiple degradations like noise, blur, and enhancing the low-light conditions.

\section{Proposed Method}
An observed image $y$, with multiple degradations can be modeled as follows,
\setlength{\belowdisplayskip}{0pt} \setlength{\belowdisplayshortskip}{0pt}
\setlength{\abovedisplayskip}{0pt} \setlength{\abovedisplayshortskip}{0pt}
\begin{equation}\label{eqn:model}
y = r \odot (k*x) + n,
\end{equation}
where $x$ is the clean image, $r$ is the irradiance map, $k$ is the blur kernel, and $n$ is the additive noise.  Here, $*$ and $\odot$ denote convolution and element wise multiplication operations, respectively.  Figure~\ref{fig:overview} gives an overview of the proposed NASFE network which consists of three task specific encoders, a fusion network and a decoder.  The deblur encoder ($E_B(.)$), denoise encoder ($E_N(.)$) and low-light encoder ($E_L(.)$) are trained to address the corresponding tasks of deblurring, denoising, and low-light enhancement, respectively. These encoders are used to extract task-specific features which are then fused using a network architecture search (NAS) based fusion block. Finally, the fused features are paseed through the decoder network to restore the face image. Additionally, with help of a classification network, we determine what degradations are present in the input image, and use them as a prior information to the task specific encoders and the fusion network. Furthermore, to improve the quality and preserve the identity in the restored face image, we extract identity information from a set of clean images corresponding to the same identity present in the degraded image  in the form of VGGFace features.  We denote them as the identity information ($I_{iden}$), and pass them as input along with the degraded image to the NASFE network to restore the face image. Besides using this identity information, we construct an identity loss $\mathcal{L}_{iden}$ to train the NASFE network.

Due to space limitations, details regarding the task specific encoders, classification network and the decoder network are provided in the supplementary document.

\subsection{Fusion Network}
\begin{figure}[htp!]
	\begin{center}
		\includegraphics[width=0.9\linewidth]{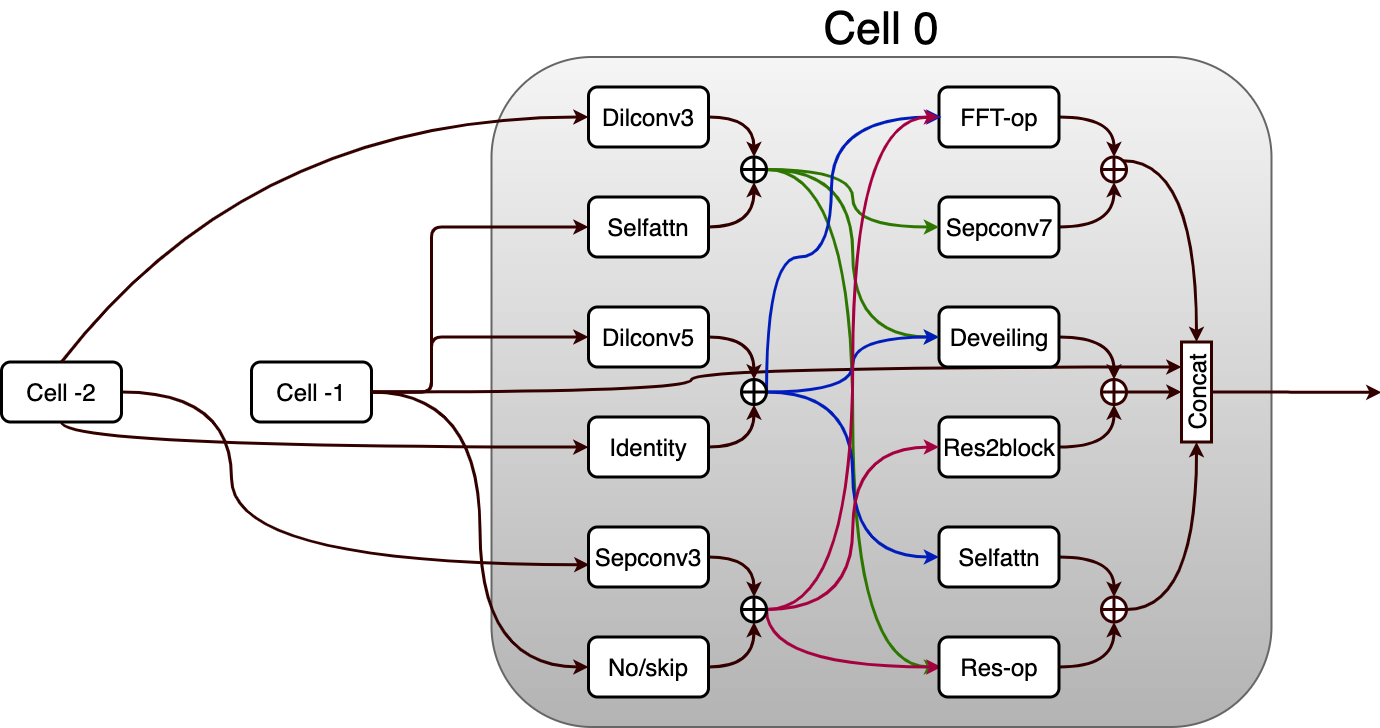}
	\end{center}
	\vskip -15pt 
	\caption{An illustration of the  Fusion cell in the fusion network of NASFE. This figure shows a possible architectural layout of the fusion cell during training.}
	\label{fig:fusion_cell}
\end{figure}

As can be seen from Fig.~\ref{fig:overview}, the task specific features are concatenated and then fed into the fusion network.  Rather than using a simple $1\times 1$ convolution or a residual block to fuse these features, we proposed to use a NAS-based approach for designing the fusion network \cite{pham2018efficient,real2019regularized,liu2018darts,liu2019auto}.  We define fusion cell as a smallest repeatable module used to construct the fusion network (see Fig.~\ref{fig:fusion_cell}).  In our approach, the network search space includes both network level search (i.e. searching the connection between different fusion cells), and cell level search (i.e. exploring structure of the fusion cell).\\

\noindent {\bf{Fusion Cell Architecture. }}
We adopt the cell design structure of ~\cite{liu2019auto} to define a fusion cell  (represented as $Cell(.)$) as a directed acyclic graph consisting of $B$ blocks. Each block $i$ in the  $l^{th}$ fusion cell $F^l$ is a two tensors to one tensor mapping structure determined as a tuple of  ($I_1,I_2,O_1,O_2,M$), where $I_1,I_2 \in \mathcal{I}_l$ are selections of the input tensors, $O_1,O_2 \in \mathcal{O}$ are selections of the layer types applied to the corresponding input tensors, and $M \in \mathcal{M}$ is the method used to combine the outputs $O_1,O_2$ to form the block $F^l_i$'s  output tensor, $Z^l_i$. The fusion cell $F^l$'s output, $Z^l$ is a concatenation of the outputs of all blocks  $\{Z^l_1,Z^l_2,\cdots,Z^l_B\}$ in the cell $F^l$.  $\mathcal{I}_l$ is the set input tensor consisting of outputs of the previous cell $F^{l-1}$ and previous-previous cell $F^{l-2}$. We follow ~\cite{liu2019auto}, and use element wise addition as the only operator of possible combining method in $\mathcal{M}$.  The set of possible layer types  $\mathcal{O}$ consist of the following ten operators:
{ \footnotesize
	\begin{itemize}[noitemsep]
		\item Dilconv3 (dilated conv$3\times3$) \inlineitem Selfattn (Self attenion)  
		\item Dilconv5 (dilated conv$5\times5$) \inlineitem Res2Block
		\item separable conv $3\times3\quad\quad\quad\;$ \inlineitem \textbf{Res-op (residual operator)}
		\item separable conv $5\times5\quad\quad\quad\;$ \inlineitem \textbf{Deveiling operator}
		\item Identity or skip connection$\quad\:$ \inlineitem \textbf{FFT operator}
		\item No or zero connection
	\end{itemize}
}

\begin{figure}[htp!]
	\begin{center}
		\includegraphics[width=\linewidth]{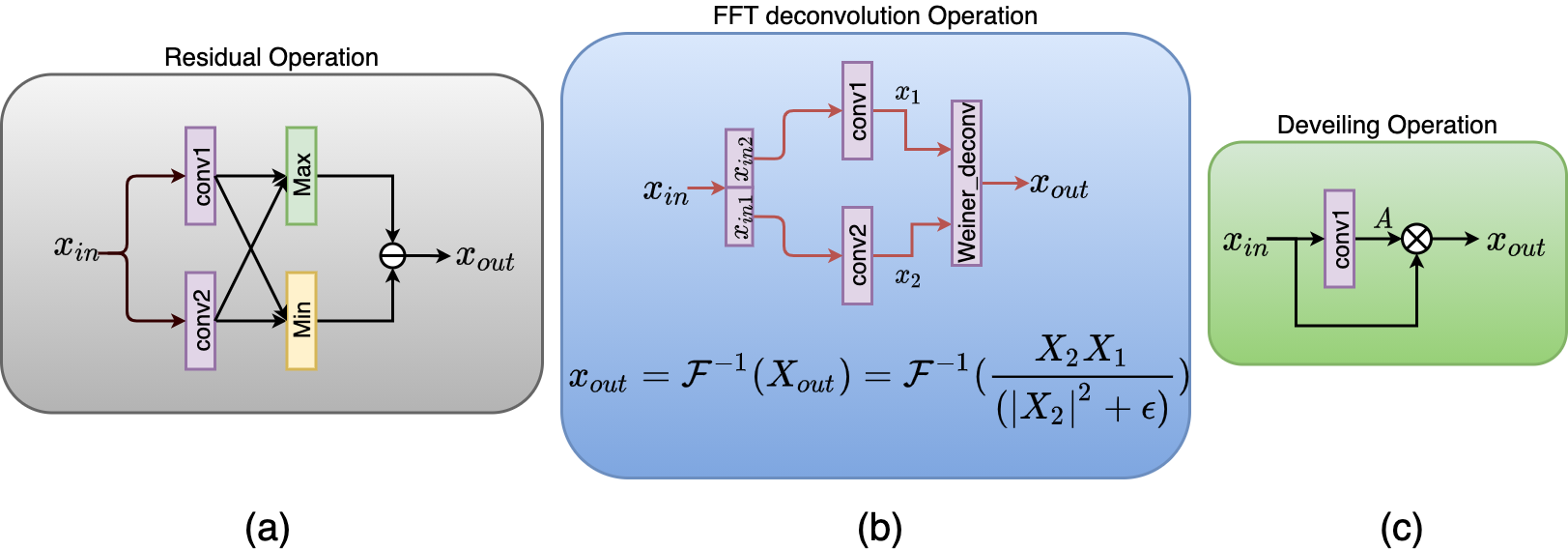}
	\end{center}
	\vskip -20pt 
	\caption{(a) Residual operator, (b) FFT-operator, (c) Deveiling operator.  All the convblocks (conv1 and conv2) in these operators are $3\times 3$  convolutions.}
	\label{fig:operators}
\end{figure}

Along with the conventional convolution operators like dilated, separable convolutions, no and skip connections, Res2Block~\cite{gao2019res2net} and self-attention block~\cite{vaswani2017attention,zhang2019self}, we introduce Res-op, deveiling and FFT-op (shown in Fig.~\ref{fig:operators}) to efficiently process the task specific features.  These operations are based on the image formation models regarding the individual degradation like  noise, blur and low-light conditions. In what follows, we explain the design of these new operators in detail.

\noindent \textbf{FFT operator. } Classical deblurring methods make use of the Wiener deconvolution technique to restore the image from blurry observations. Motivated by Wiener deconvolution, we split the features $x_{in}$ into two parts $x_{in1}$ and $x_{in2}$, and then apply convolution operation to obtain $x_1$ and $x_2$, respectively. Finally, as shown in  Fig.~\ref{fig:operators}(b), we apply Weiner deconvolution to obtain $x_{out}=\mathcal{F}^{-1}(X_{out})$, where 
\begin{equation}
X_{out}=\frac{X_2X_1}{(|X_2|^2+\epsilon)}.
\end{equation} 
Here,  $\mathcal{F}^{-1}$ denotes the inverse Fourier transform operator and $X_1$, $X_2$ and $X_{out}$ are the Fourier transforms of $x_1$, $x_2$ and $x_{out}$, respectively. Here, $\epsilon$ resembles the inverse of signal-to-noise ratio used during Weiner deconvolution which we set equal to 0.01.\\

\noindent \textbf{Deveiling operator. } \cite{Dono2005,PatchBasedHDR} viewed low-light enhancement as an image dehazing problem since both have similar mathematical models. \cite{Dono2005} addressed  low-light enhancement problem using a dark channel prior. Motivated by these methods, we use a deveiling operator \cite{li2017aod,li2019rainflow} to learn a latent mask $A$ that enhances features from low-light conditions as follows
\begin{equation}
	y = r \odot x + n
	\implies 
	x = A\odot y, 
\end{equation}
where $A =  \frac{(y-n) }{r y}$ is a learnable mask and  a function of $y$. As shown in Fig.~\ref{fig:operators}(c) $A$ can be learned using a convolution layer given input features $x_{in}$ and element wise multiplication of $A$ with $x_{in}$ to obtain the enhanced features $x_{out}$.\\

\noindent \textbf{Residual operator.} Inspired by the image deraining work \cite{li2018robust} that uses a residual operator to remove rain streaks, we use a similar residual operator shown in Fig.~\ref{fig:operators}(a) to estimate noise from the latent features and finally obtain noisy free features.\\

\noindent {\bf{Fusion Cell search space. }}
We use the continuous relaxation approach ~\cite{liu2017hierarchical,liu2018darts} in which every output tensor $Z_i^l$ of block $F_i^l$  is connected to all input tensors $I_l^i$ through operation $O_{j\rightarrow i}$ as follows
\begin{equation}\label{eqn:arch1}
Z_i^l = \sum_{Z^l_j\in I^l_i} O_{j\rightarrow i}\left(Z^l_j\right).
\end{equation}
We define  $\bar O_{j\rightarrow i}$, as approximation of the best search for operators $O_{j\rightarrow i}$ using  continuous relaxation as follows
\begin{equation}\label{eqn:arch2}
\bar{O}_{j \rightarrow i}\left(Z_{j}^{l}\right)=\sum_{O^{k} \in \mathcal{O}} \alpha_{j \rightarrow i}^{k} O^{k}\left(Z_{j}^{l}\right),
\end{equation}
where $\sum_{k=1}^{|\mathcal{O}|} \alpha_{j \rightarrow i}^{k}=1$, and  $\alpha_{j \rightarrow i}^{k} \geq 0 \quad \forall i, j$. This can be easily implemented using softmax. Hence, using Eq.~\ref{eqn:arch1} and Eq.~\ref{eqn:arch2}, we can summarize the fusion cell architecture as follows,
\begin{equation}
Z^l = Cell(Z^{l-1},Z^{l-2},\alpha).
\end{equation}

\subsection{Identity Information}\label{sec:iden}

Given the degraded image $y$, and a set of clean images  $\mathcal{D}_C=\{C_i\}_{i=1}^{n}$, we compute pool3 features using the VGGFace network~\cite{Parkhi15}.  Note that $\mathcal{D}_C$ contains clean images of the same person present in the degraded image $y$.  Let $F^y$ and $\{F^C_i\}_{i=1}^{n}$ denote the VGGFace features corresponding to $y$ and  $\{C_i\}_{i=1}^{n}$, respectively.  Since the clean images in $\mathcal{D}_C$ may have different style and characteristics as they may have been taken in different scenarios and times, we apply Adaptive-Instance normalization (AdaIN)  before passing them as input to the network to reduce the effect of different styles in the images.  AdaIN is applied as follows, 
\begin{equation}\label{eqn:adain}
\bar F^C_i = \sigma(F^y)\left(\frac{F^C_i - \mu(F^C_i)}{\sigma(F^C_i)}\right) +\mu(F^y),
\end{equation}
where $\sigma(.)$ and $\mu(.)$ denote standard deviation and mean, respectively. Mean of $\bar F^C_i $ is defined as the identity information, i.e. $I_{iden} = mean(\{\bar F^C_i\}) $.  $I_{iden}$ is used along with $y$ as input to the NASFE network to restore the face image.

Note that as we are extracting the identity information from the clean images, this information is much more reliable and provides stronger prior as compared to face exemplar masks \cite{PanECCV14}  or semantic maps~\cite{ZiyiDeep,bulat2018super,yasarla2020deblurring} extracted using the degraded images.  Additionally, to preserve the identity of the subject in restored image, we construct an identity loss $\mathcal{L}_{iden}$ to train the NASFE network.\\

\noindent \textbf{Identity Loss $\mathcal{L}_{iden}$.}    Let $\hat{x}$ denote the restored face image using the NASFE network. We construct the identity loss as follows
\begin{equation}\label{eq:idloss}
\mathcal{L}_{iden} = \frac{1}{n}\sum_{i=1}^{n}  \arccos(\langle F^{\hat{x}},F^C_i\rangle ),
\end{equation}
where $F^{\hat{x}}$ denotes the VGGFace features corresponding to $\hat{x}$ and $n$ denotes the number of clean images in $\mathcal{D}_C$.

\subsection{Overall Loss}
The NASFE network is trained using a combination of the L2 loss, perceptual loss \cite{johnson2016perceptual} and identity loss as follows
\begin{equation}
\mathcal{L}_{final} = \mathcal{L}_{2} + \lambda_{per}  \mathcal{L}_{per}  + \lambda_{iden} \mathcal{L}_{iden}
\end{equation}
where $\mathcal{L}_{2} = \|\hat{x} -x\|_2^2$,  $\mathcal{L}_{iden} $ denotes the identity loss defined in \eqref{eq:idloss} and  $\mathcal{L}_{per}$ denotes the perceptual loss~\cite{johnson2016perceptual} defined as follows
\begin{equation}
\mathcal{L}_{per} = \frac{1}{NHW} \sum_{i}\sum_{j}\sum_{k} \| F^{\hat x} _{i,j,k}-F^{x}_{i,j,k}\|.
\end{equation} 
Here, $F^{\hat x}$, $F^{x}$ denote the \textit{pool3} layer features of the VGGFace network~\cite{Parkhi15} and $N, H$ and $W$ are the number of channels, height and width of $F^{\hat x}$, respectively.  We set  $\lambda_{per}=0.04$ and $ \lambda_{iden}=0.003$ in our experiments.

Note that multiple clean images are required only during training.  Once the network is trained, a degraded image is fed into the network and the NASFE produces identity-preserving restored image as the output.


\begin{table*}[h!]
	\caption{PSNR$|$SSIM comparisons against SOTA denoising methods using \textit{Test-N}}
	\vskip-10pt
	\label{table:noise_comp}
	\setlength{\tabcolsep}{8pt} 
	\renewcommand{\arraystretch}{1.25} 
	\resizebox{0.9\textwidth}{!}{
		\begin{tabular}{l|l|cccccc|c}
			\hline
			\multicolumn{2}{l|}{Test-set}     & Noisy        & \begin{tabular}[c]{@{}l@{}}BM3D~\cite{dabov2007image} \\ (TIP'7)  \end{tabular}       & \begin{tabular}[c]{@{}l@{}} EPLL~\cite{zoran2011learning}   \\ (ICCV'11)   \end{tabular}   & \begin{tabular}[c]{@{}l@{}} TNRD~\cite{chen2016trainable}  \\ (PAMI'16)  \end{tabular}     & \begin{tabular}[c]{@{}l@{}} DnCNN~\cite{zhang2017beyond} \\\ (TIP'17)  \end{tabular}     & \begin{tabular}[c]{@{}l@{}} CBDNet~\cite{guo2019toward}  \\ (CVPR'19)  \end{tabular}   & \begin{tabular}[c]{@{}l@{}} NASFE \\ (ours) \end{tabular} \\ \hline
			\multirow{2}{*}{\textit{Test-N}} & CelebA   & 17.63$|$0.50 & 24.50$|$0.75 & 23.59$|$0.71 & 26.36$|$0.79 & 27.54$|$0.87 & 28.74$|$0.88 & \textbf{29.23$|$0.90} \\ \cline{2-9} 
			& VGGface2 & 18.52$|$0.46 & 24.45$|$0.73 & 23.87$|$0.73 & 26.91$|$0.81 & 27.71$|$0.88 & 28.88$|$0.89 & \textbf{29.56$|$0.91} \\ \hline
		\end{tabular}
	}
\end{table*}


\begin{table*}[h!]
	\caption{PSNR$|$SSIM comparisons against SOTA deblurring methods using \textit{Test-B}}
	\vskip-10pt
	\label{table:blur_comp}
	\setlength{\tabcolsep}{8pt} 
	\renewcommand{\arraystretch}{1.25} 
	\resizebox{0.9\textwidth}{!}{
		\begin{tabular}{l|l|ccccccc|c}
			\hline
			\multicolumn{2}{l|}{Test-set} & Blurry & \begin{tabular}[c]{@{}l@{}} Pan \etal~\cite{PanECCV14} \\ (ECCV'14) \end{tabular}& \begin{tabular}[c]{@{}l@{}} Super-FAN~\cite{bulat2018super} \\ (CVPR'18) \end{tabular} & \begin{tabular}[c]{@{}l@{}} Shen \etal~\cite{ZiyiDeep} \\ (CVPR'19) \end{tabular} & \begin{tabular}[c]{@{}l@{}} UMSN~\cite{yasarla2020deblurring} \\ (TIP'20) \end{tabular} & \begin{tabular}[c]{@{}l@{}} DebluGAN-v2~\cite{kupyn2019deblurgan} \\ (ICCV'19) \end{tabular}  & \begin{tabular}[c]{@{}l@{}} HiFaceGAN~\cite{Li_2020_ECCV} \\ (ACMM'20) \end{tabular} & \begin{tabular}[c]{@{}l@{}}NASFE\\ (ours)\end{tabular} \\ \hline
			\multirow{2}{*}{\textit{Test-B}} & CelebA & 19.64$|$0.61 & 19.98$|$0.74 &  22.98$|$0.80 & 24.45$|$0.83 & 25.53$|$0.87 & 26.06$|$0.88 & 26.38$|$0.88 & \textbf{27.90$|$0.92} \\ \cline{2-2}
			& VGGface2 & 20.30$|$0.67 & 20.74$|$0.77  & 22.83$|$0.80 & 24.63$|$0.84 & 25.65$|$0.87 & 25.92$|$0.87  & 26.30$|$0.88 & \textbf{27.94$|$0.91} \\ \hline
		\end{tabular}
	}
\end{table*}

\begin{table*}[h!]
	\caption{PSNR$|$SSIM comparisons against SOTA low-light enhancement methods using \textit{Test-L}}
	\vskip-10pt
	\label{table:low_comp}
	\setlength{\tabcolsep}{8pt} 
	\renewcommand{\arraystretch}{1.25} 
	\resizebox{0.9\textwidth}{!}{
		\begin{tabular}{l|l|cccccc|c}
			\hline
			\multicolumn{2}{l|}{Test set} & Low-light & \begin{tabular}[c]{@{}c@{}}Fu\etal\cite{fu2016weighted}\\ +DnCNN\cite{zhang2017beyond}\end{tabular} & \begin{tabular}[c]{@{}c@{}}LIME\cite{guo2016lime}\\ +DnCNN\cite{zhang2017beyond}\end{tabular} & \begin{tabular}[c]{@{}c@{}}Li\etal\cite{li2018structure}\\ +DnCNN\cite{zhang2017beyond}\end{tabular} & \begin{tabular}[c]{@{}c@{}}Zero-DCE\cite{guo2020zero}\\ +DnCNN\cite{zhang2017beyond}\end{tabular} & \begin{tabular}[c]{@{}c@{}}Zero-DCE\cite{guo2020zero}+\\ ELD~\cite{wei2020physics}\end{tabular} & \begin{tabular}[c]{@{}l@{}}NASFE\\ Ours\end{tabular} \\ \hline
			\multirow{2}{*}{Test-N} & CelebA & 8.25$|$0.14 & 16.28$|$0.44 & 17.18$|$0.49 & 18.44$|$0.57 & 19.85$|$0.67 & 20.15$|$0.71 & \textbf{23.04$|$0.80} \\ \cline{2-9} 
			& VGGFace2 & 8.09$|$0.12 & 15.83$|$0.40 & 16.72$|$0.47 & 17.79$|$0.54 & 19.29$|$0.64 & 19.83$|$0.67 & \textbf{22.94$|$0.79} \\ \hline
		\end{tabular}
	}
\end{table*} 

\section{NASFE Implementation Details}
Given $x$, we first convolve it with $k$ to get a blurry image.  Here, $k$ can be a motion blur kernel~\cite{boracchi2012modeling,kupyn2018deblurgan} or an anisotropic Gaussian blur kernel~\cite{zhang2018learning}.  To generate a degraded image with blur+low light+noise conditions, we follow~\cite{guo2019toward,wei2020physics} and convert the obtained blurry image to irradiance image $L$.   We then multiply low light factor $r$ to $L$, where $L = Mf^{-1}(k*x)$, $f(.)$ is the camera response function (CRF) function, and $M(.)$ represents the function that converts an RGB image to a Bayer image. Finally, we add realistic Photon-Gaussian noise~\cite{guo2019toward}, where $n$ consists of two components: stationary noise $n_c$ with noise variance $\sigma^2_c$ and signal dependent noise $n_s$ with spatially varying noise variance $L\sigma_s^2$.

\noindent \textbf{Training dataset.} We conduct our experiments using clean images from the CelebA ~\cite{liu2015faceattributes} and VGGFace2~\cite{cao2018vggface2} face datasets. We randomly selected 30000 images from the  training set of CelebA ~\cite{liu2015faceattributes}, and 30000 images from VGGFace2~\cite{cao2018vggface2} and generate synthetic degraded images with multiple degradations. Images in the CelebA and VGGFace2 datasets are of size $176 \times 144$ and $224 \times 224$, respectively.  Given a clean face image $x$, we first convolve it with blur kernel $k$ sampled from 25000 motion kernels generated using ~\cite{boracchi2012modeling,kupyn2018deblurgan}, and 8 anisotoric Gaussian kernels~\cite{zhang2018learning}, and then following \cite{wei2020physics,xu2020learning} we multiply them  with low light factor ($r$) sampled uniformly from [0.05, 0.5] to obtain images with low-light conditions. Finally, we add realistic noise~\cite{guo2019toward} $n$ (where $\sigma_s \in [0.01,0.16]$ and $\sigma_c \in [0.01,0.06]$) to obtain the degraded image $y$. Based on the degradations present in $y$, we create class label $c$ which is a vector of length three, $c= \{b,n,l\}$ where $b, n$, and $l$ are binary numbers, \textit{i.e} $b, n$, and $l$ are one if $y$ contains  blur, noise and low-light, respectively and zero otherwise.

\noindent \textbf{Test datasets.} We create test datasets using randomly sampled 100 test images from the test sets of  CelebA~\cite{liu2015faceattributes}  and VGGFace2~\cite{cao2018vggface2}. Using these clean images, we create test datasets \textit{Test-B, Test-N, Test-L, Test-BN,} and \textit{Test-BNL} with the amounts of degradations as shown in the Table~\ref{table:test_data}. Additionally, we collected a real-world face image dataset with multiple degradations corresponding to 20 subjects from YouTube.

\noindent {\bf{Training Details.}}
The NASFE network is trained using 
Given $\{y_{i}, x_{i}, c_{i}, \mathcal{D}_{C}^{i}\}_{i=1}^{N}$.  The classification network (CN) is trained using $\{y_{i},c_{i}\}_{i=1}^{N}$.  It is trained to produce a class label $\hat{c}_{i}$ which indicates degradation(s) present in $y_{i}$. 

\begin{table}[h!]
	\caption{Details of the test datasets created using CelebA~\cite{Parkhi15} and VGG-Face2~\cite{cao2018vggface2}. M: motion kernels~\cite{boracchi2012modeling}, Gaussian: Gaussian kernels~\cite{zhang2018learning}}
	\vskip-10pt
	\label{table:test_data}
	\setlength{\tabcolsep}{8pt} 
	\renewcommand{\arraystretch}{1.25} 
	\resizebox{1.0\textwidth}{!}{
		\begin{tabular}{l|c|c|cc}
			\hline
			\multirow{2}{*}{Test set name} & \multirow{2}{*}{\begin{tabular}[c]{@{}l@{}}Degradation\\ type\end{tabular}} & \multirow{2}{*}{\begin{tabular}[c]{@{}c@{}}Details about\\ degradation values\end{tabular}} & \multicolumn{2}{c}{Number of images} \\ \cline{4-5} 
			&  &  & CelebA & VGGFace2 \\ \hline
			\textit{Test-B} & blur & \begin{tabular}[c]{@{}c@{}}M: 40 kernels kernel size $[13,27]$ \\ G: 12 anisotropic kernels with $\sigma_b\in[1,3]$\end{tabular} & 5200 & 5200 \\ \hline
			\textit{Test-N} & noise & \begin{tabular}[c]{@{}c@{}}$\sigma_s = [0.05,0.1,0.05]$, $\sigma_c = [0.05,0.1]$\end{tabular} & 600 & 600 \\ \hline
			\textit{Test-L} & low-light & \begin{tabular}[c]{@{}c@{}}$r = [0.1,0.15,0.2,0.25,0.3,0.35]$ \\ $\sigma_s = [0.05,0.1]$, $\sigma_c = [0.05,0.1]$\end{tabular} & 2400 & 2400 \\ \hline
			\textit{Test-BN} & blur + noise & \begin{tabular}[c]{@{}c@{}}M: 40 kernels kernel size $[13,27]$ \\ G: 12 anisotropic kernels with $\sigma_b\in[1,3]$\\ $\sigma_s = 0.1, \sigma_c=0.05$\end{tabular} & 10400 & 10400 \\ \hline
			\textit{Test-BNL} & \begin{tabular}[c]{@{}c@{}} blur + noise \\ + low-light \end{tabular}& \begin{tabular}[c]{@{}c@{}}M: 40 kernels kernel size $[13,27]$ \\ G: 12 anisotropic kernels with $\sigma_b\in[1,3]$\\ $r = [0.15,0.3], \sigma_s = 0.1, \sigma_c=0.05$\end{tabular} & 10400 & 10400 \\ \hline
		\end{tabular}
	}
\end{table}

 Training and network details of the classification network are provided in the supplementary document. NASFE contains  three encoders (deblur ($E_B(.)$), denoise ($E_N(.)$) and low-light ($E_L(.)$)), one fusion network ($Fn(.)$) and a decoder ($De(.)$). Encoders ($E_B,E_N,E_L$) are initially trained to address the corresponding individual tasks of deblurring, denoising, and low-light enhancement, respectively. More details are provided in the supplementary document. Given a degraded image $y$, we compute class $\hat{c}$ (using CN) and identity information $I_{iden}$ and pass them as input to NASFE to compute a restored image $\hat{x}$. We set the number of blocks $B$ to 12 in the Fusion cell of NASFE.  Following ~\cite{liu2019auto}, we update $\alpha$ and the weights of the NASFE  alternately during training. NASFE is trained using $\mathcal{L}_{final}$ with the Adam optimizer and batch size of 40. The learning rate is set equal to 0.0005.  NASFE is trained for one million iterations.

\begin{figure*}[htp!]
	\centering
	\includegraphics[width=0.1\textwidth,height=0.13\textwidth]{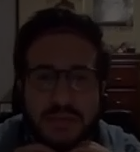}
	\includegraphics[width=0.1\textwidth,height=0.13\textwidth]{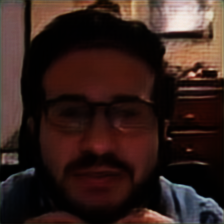}
	\includegraphics[width=0.1\textwidth,height=0.13\textwidth]{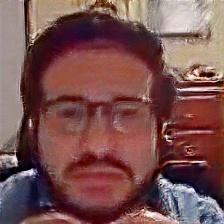}
	\includegraphics[width=0.1\textwidth,height=0.13\textwidth]{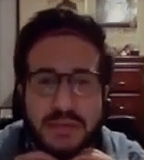}
	\includegraphics[width=0.1\textwidth,height=0.13\textwidth]{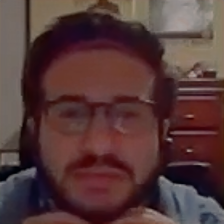}
	\includegraphics[width=0.1\textwidth,height=0.13\textwidth]{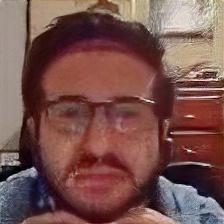}
	\includegraphics[width=0.1\textwidth,height=0.13\textwidth]{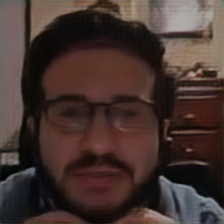}
	\includegraphics[width=0.1\textwidth,height=0.13\textwidth]{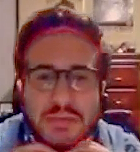}
	\includegraphics[width=0.1\textwidth,height=0.13\textwidth]{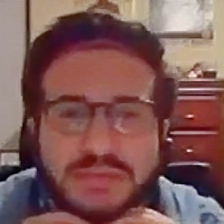}
	\\ \vskip1pt
	\includegraphics[width=0.1\textwidth,height=0.13\textwidth]{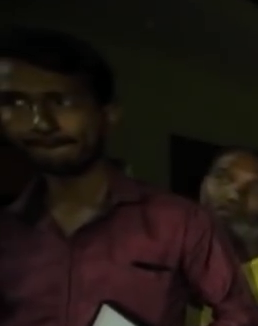}
	\includegraphics[width=0.1\textwidth,height=0.13\textwidth]{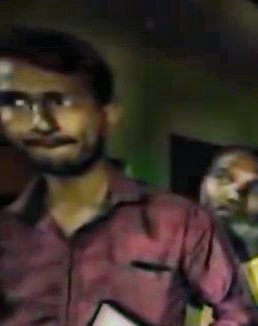}
	\includegraphics[width=0.1\textwidth,height=0.13\textwidth]{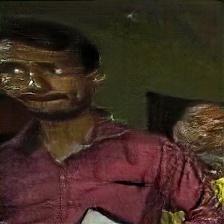}
	\includegraphics[width=0.1\textwidth,height=0.13\textwidth]{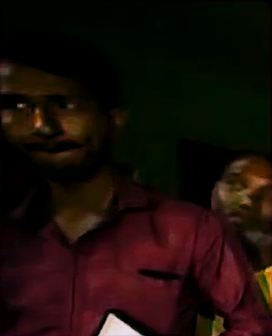}
	\includegraphics[width=0.1\textwidth,height=0.13\textwidth]{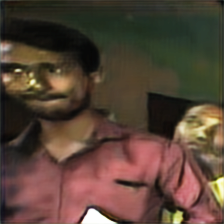}
	\includegraphics[width=0.1\textwidth,height=0.13\textwidth]{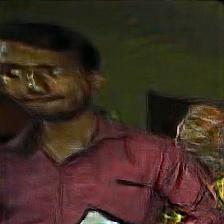}
	\includegraphics[width=0.1\textwidth,height=0.13\textwidth]{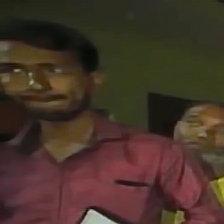}
	\includegraphics[width=0.1\textwidth,height=0.13\textwidth]{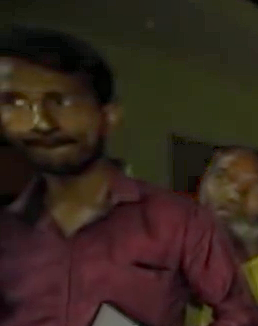}
	\includegraphics[width=0.1\textwidth,height=0.13\textwidth]{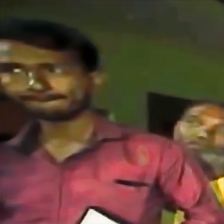}
	\\ \vskip1pt
	\includegraphics[width=0.1\textwidth,height=0.13\textwidth]{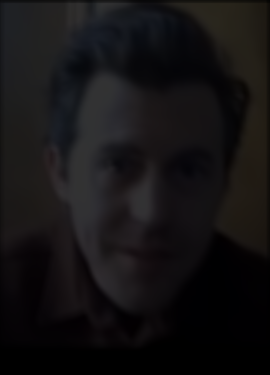}
	\includegraphics[width=0.1\textwidth,height=0.13\textwidth]{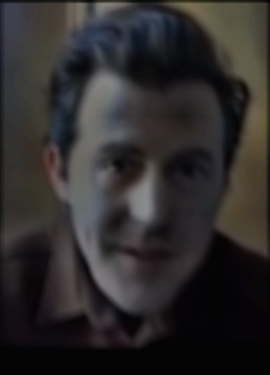}
	\includegraphics[width=0.1\textwidth,height=0.13\textwidth]{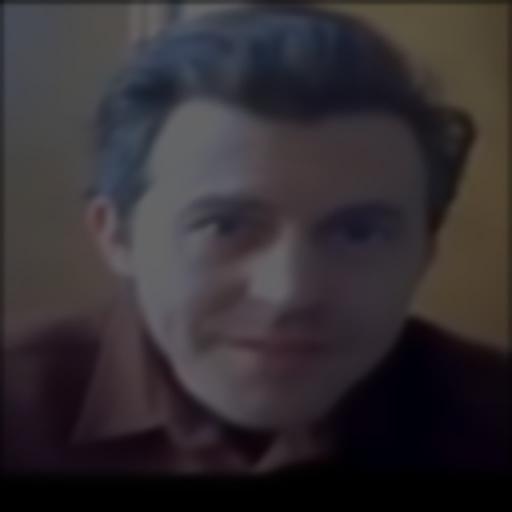}
	\includegraphics[width=0.1\textwidth,height=0.13\textwidth]{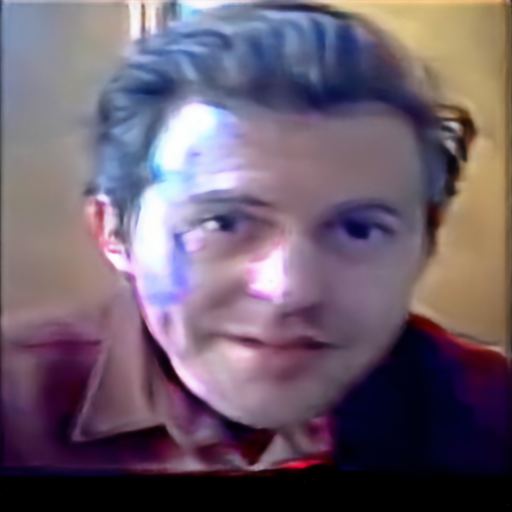}
	\includegraphics[width=0.1\textwidth,height=0.13\textwidth]{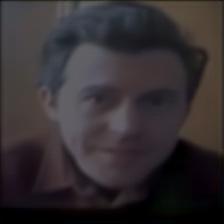}
	\includegraphics[width=0.1\textwidth,height=0.13\textwidth]{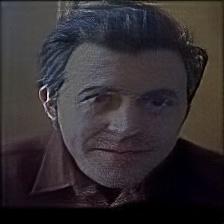}
	\includegraphics[width=0.1\textwidth,height=0.13\textwidth]{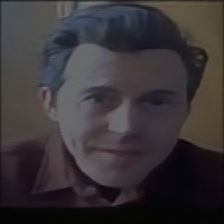}
	\includegraphics[width=0.1\textwidth,height=0.13\textwidth]{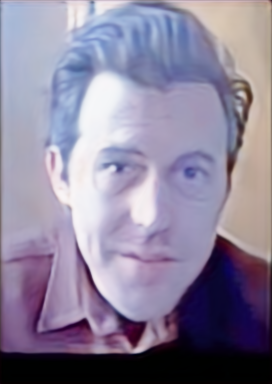}
	\includegraphics[width=0.1\textwidth,height=0.13\textwidth]{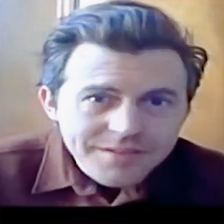}
	\\
	\begin{flushleft}
		\vskip-15pt
		{\scriptsize \hskip28pt Input image \hskip20pt Histeq \hskip20pt Super-FAN~\cite{bulat2018super} \hskip10pt Shen~\textit{et al.}~\cite{ZiyiDeep} \hskip14pt UMSN~\cite{yasarla2020deblurring} \hskip6pt DeblurGANv2\cite{kupyn2019deblurgan}  \hskip6pt DFDNet~\cite{Li_2020_ECCV} \hskip3pt HiFaceGAN~\cite{Yang2020HiFaceGANFR} \hskip16pt NASFE }\\ 
	\end{flushleft}
	\vskip-20pt
	\caption{Qualitative comparisons using real face images multiple degradations collected from YouTube videos. output images of Histeq are computed using Histogram equalization method.}
	\label{Fig:qual_BNL}
\end{figure*}
\begin{figure*}[htp!]
	\centering
	\includegraphics[width=0.1\textwidth,height=0.13\textwidth]{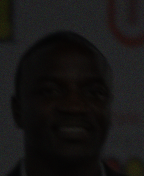}
	\includegraphics[width=0.1\textwidth,height=0.13\textwidth]{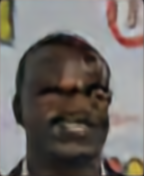}
	\includegraphics[width=0.1\textwidth,height=0.13\textwidth]{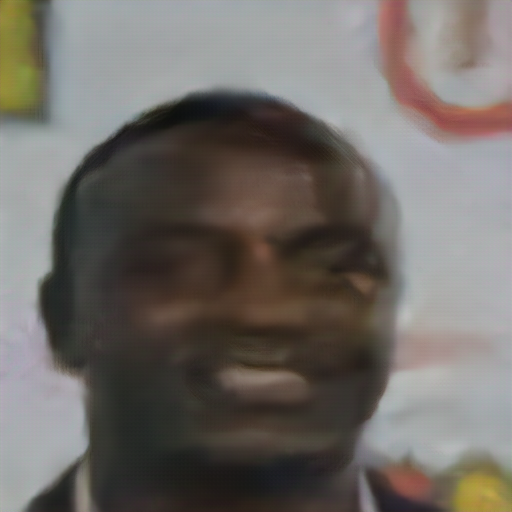}
	\includegraphics[width=0.1\textwidth,height=0.13\textwidth]{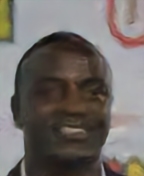}
	\includegraphics[width=0.1\textwidth,height=0.13\textwidth]{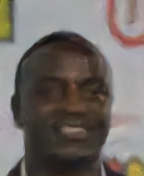}
	\includegraphics[width=0.1\textwidth,height=0.13\textwidth]{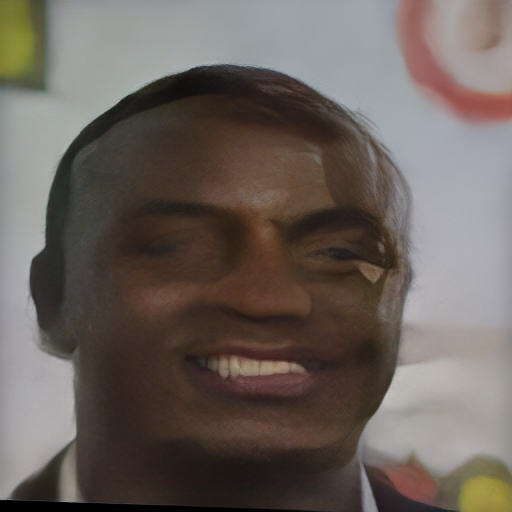}
	\includegraphics[width=0.1\textwidth,height=0.13\textwidth]{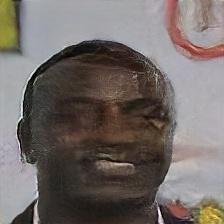}
	\includegraphics[width=0.1\textwidth,height=0.13\textwidth]{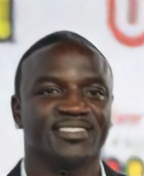}
	\includegraphics[width=0.1\textwidth,height=0.13\textwidth]{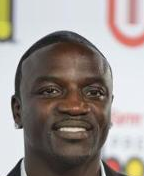}
	\\ \vskip1pt
	\includegraphics[width=0.1\textwidth,height=0.13\textwidth]{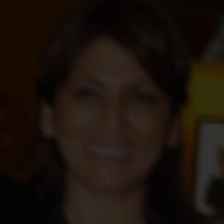}
	\includegraphics[width=0.1\textwidth,height=0.13\textwidth]{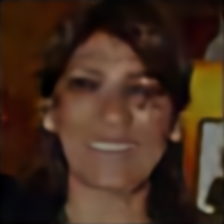}
	\includegraphics[width=0.1\textwidth,height=0.13\textwidth]{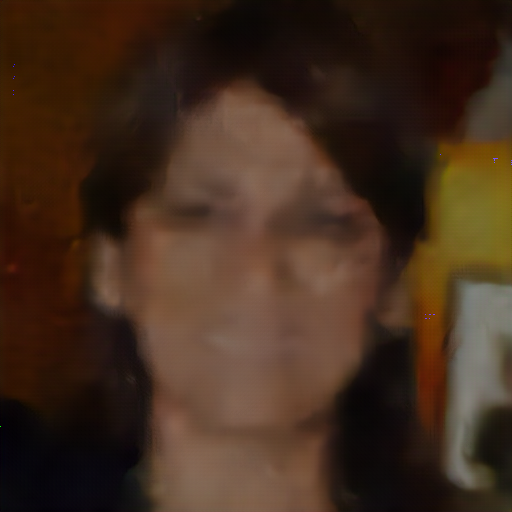}
	\includegraphics[width=0.1\textwidth,height=0.13\textwidth]{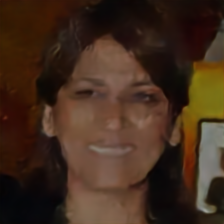}
	\includegraphics[width=0.1\textwidth,height=0.13\textwidth]{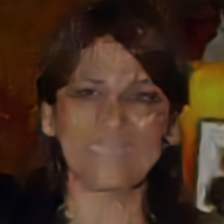}
	\includegraphics[width=0.1\textwidth,height=0.13\textwidth]{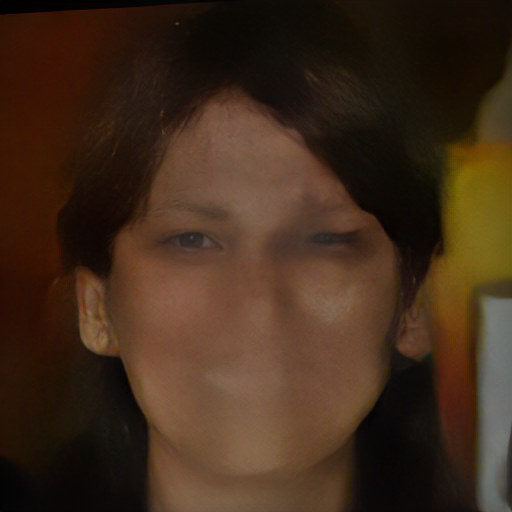}
	\includegraphics[width=0.1\textwidth,height=0.13\textwidth]{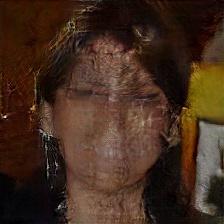}
	\includegraphics[width=0.1\textwidth,height=0.13\textwidth]{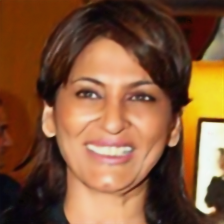}
	\includegraphics[width=0.1\textwidth,height=0.13\textwidth]{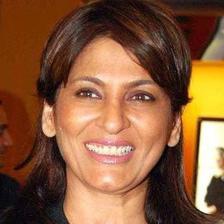}
	\\ \vskip1pt
	\includegraphics[width=0.1\textwidth,height=0.13\textwidth]{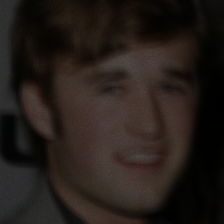}
	\includegraphics[width=0.1\textwidth,height=0.13\textwidth]{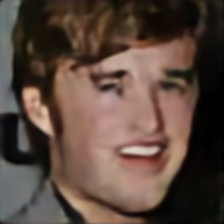}
	\includegraphics[width=0.1\textwidth,height=0.13\textwidth]{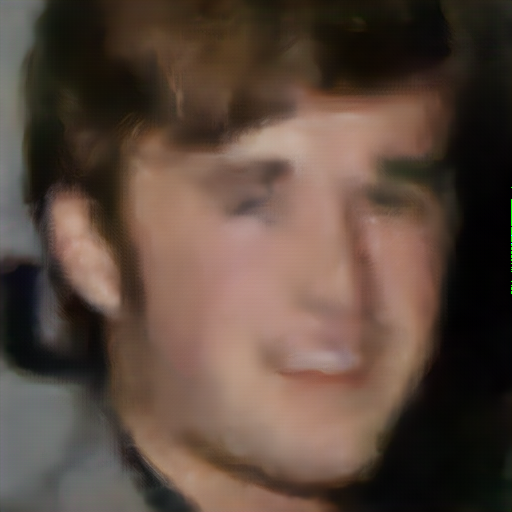}
	\includegraphics[width=0.1\textwidth,height=0.13\textwidth]{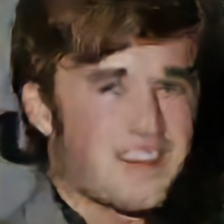}
	\includegraphics[width=0.1\textwidth,height=0.13\textwidth]{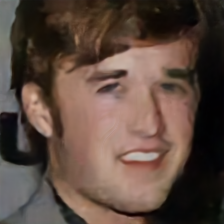}
	\includegraphics[width=0.1\textwidth,height=0.13\textwidth]{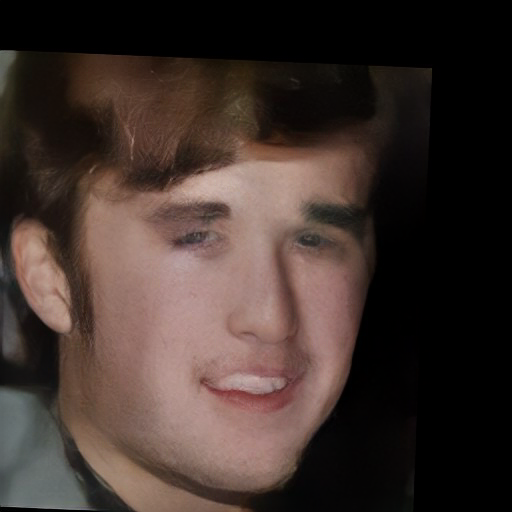}
	\includegraphics[width=0.1\textwidth,height=0.13\textwidth]{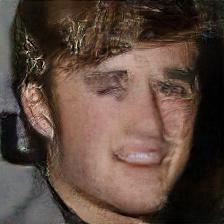}
	\includegraphics[width=0.1\textwidth,height=0.13\textwidth]{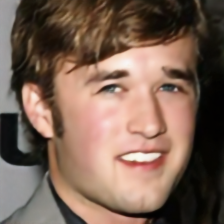}
	\includegraphics[width=0.1\textwidth,height=0.13\textwidth]{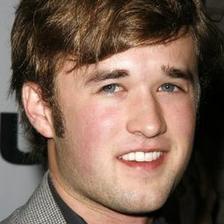}
	\\
	\begin{flushleft}
		\vskip-15pt
		{\scriptsize \hskip30pt B+N+L \hskip16pt Shen~\textit{et al.}~\cite{ZiyiDeep}\hskip10pt Super-FAN~\cite{bulat2018super}  \hskip14pt UMSN~\cite{yasarla2020deblurring} \hskip6pt DeblurGANv2\cite{kupyn2019deblurgan}  \hskip6pt DFDNet~\cite{Li_2020_ECCV} \hskip5pt HiFaceGAN~\cite{Yang2020HiFaceGANFR} \hskip16pt NASFE \hskip22pt Ground-Truth}\\ 
	\end{flushleft}
	\vskip-20pt
	\caption{Qualitative comparisons using synthetic test set \textit{Test-BNL}, B:blur, N: noise, L: low-light condition.}
	\label{Fig:qual_BNL_syn}
\end{figure*}
\begin{table*}[htp!]
	\caption{PSNR$|$SSIM and face recognition comparisons of NASFE using \textit{Test-BNL}. B: blur, N: noise, L: low-light. (Note all methods are retrained using degraded images containing blur, noise and low-light conditions.  Histeq means Histogram equalization method)}
	\vskip-10pt
	\label{table:blur+noise+low_comp}
	\setlength{\tabcolsep}{8pt} 
	\renewcommand{\arraystretch}{1.25} 
	\resizebox{\textwidth}{!}{
		\begin{tabular}{l|l|l|cccccccc|cc}
			\hline
			\multicolumn{2}{l|}{Test-set} & Metrics & B+N+L  & Histeq & \begin{tabular}[c]{@{}c@{}}Shen \etal~\cite{ZiyiDeep}\\ (CVPR'18)\end{tabular} & \multicolumn{1}{c}{\begin{tabular}[c]{@{}c@{}}Super-FAN~\cite{bulat2018super}\\ (CVPR'18)\end{tabular}} &  \multicolumn{1}{c}{\begin{tabular}[c]{@{}c@{}}UMSN~\cite{yasarla2020deblurring}\\ (TIP'20)\end{tabular}} & \begin{tabular}[c]{@{}c@{}}DeblurGANv2~\cite{kupyn2019deblurgan}\\ (ICCV'19)\end{tabular} &  \begin{tabular}[c]{@{}c@{}}DFDNet~\cite{kupyn2019deblurgan}\\ (ECCV'20)\end{tabular} &  \begin{tabular}[c]{@{}c@{}}HiFaceGAN~\cite{kupyn2019deblurgan}\\ (ACMM'20)\end{tabular} &\begin{tabular}[c]{@{}c@{}}NASFE \\ w/o $I_{iden}$\end{tabular} & \begin{tabular}[c]{@{}c@{}}NASFE \\ (ours)\end{tabular}\\ \hline
			\multirow{4}{*}{Test-BNL} & \multirow{2}{*}{CelebA} & PSNR$|$SSIM &8.17$|$0.22 &12.28$|$0.39 &  19.98$|$0.66&  19.45$|$0.60&  21.10$|$ 0.72&  21.37$|$0.74& 21.68$|$0.77 & 21.75$|$0.77& 22.53$|$0.81 &\textbf{23.80$|$0.85} \\
			&  & Top-1$|$Top-5 & 38.5$|$45.2 & 41.3$|$47.5 &  51.2$|$57.8 &  48.8$|$55.9 &  59.7$|$67.7 &   61.2 $|$ 68.6  & 63.4$|$69.7& 62.9$|$69.1 & 69.6$|$77.3 &\textbf{73.4$|$81.8}\\ \cline{2-13} 
			& \multirow{2}{*}{VGGFace2} & PSNR$|$SSIM & 8.39$|$0.25 & 12.84$|$0.41&  20.46$|$0.68 &  19.83$|$0.64 &21.28$|$ 0.72 &  21.94$|$0.78 & 21.96$|$0.78 & 22.08$|$0.79 & 22.84$|$0.83 &\textbf{24.15$|$0.86} \\
			& & Top-1$|$Top-5 & 42.8$|$49.2 & 43.2$|$50.8   &  56.9$|$62.7 &  53.8$|$59.2 &  62.8$|$ 69.9 &  65.3$|$72.4 & 67.8$|$73.7 & 67.2$|$72.9 & 72.1$|$80.4 & \textbf{75.9$|$84.2}\\ \hline
		\end{tabular}
	}
\end{table*}

\section{Experiments and Results}
We compare the performance of our network against the state-of-the-art (SOTA) denoising ~\cite{dabov2007image,zoran2011learning,chen2016trainable,zhang2017beyond,guo2019toward}, deblurring~\cite{PanECCV14,bulat2018super,ZiyiDeep,yasarla2020deblurring,kupyn2019deblurgan,Li_2020_ECCV,Yang2020HiFaceGANFR}, and low-light enhancement methods~\cite{fu2016weighted,guo2016lime,li2018structure,guo2020zero,wei2020physics}. Peak-Signal-to-Noise Ratio (PSNR) and Structural Similarity index (SSIM)~\cite{wang2004image} measures are used to compare the performance of different methods on synthetic images. Note that we retrain the SOTA methods \cite{bulat2018super,ZiyiDeep,yasarla2020deblurring,kupyn2019deblurgan,Li_2020_ECCV,Yang2020HiFaceGANFR} using the training data discussed earlier and following the procedures explained in the respective papers. Additionally, we provide visual comparisons of NASFE against the SOTA methods using the real blurry images provided by~\cite{ZiyiDeep}, and real low-light exposure images published by~\cite{guo2020zero,cai2017efficient}.

\subsection{Single degradation experiments}
\noindent {\bf{Denoising Experiments.}}  We perform quantitative analysis of NASFE against the SOTA  denoising methods~\cite{dabov2007image,zoran2011learning,chen2016trainable,zhang2017beyond,guo2019toward} using \textit{Test-N}. As  can be seen from Table~\ref{table:noise_comp}, our method  outperformed the SOTA method by 0.6dB in PSNR and 0.02 in SSIM. Note that methods such as \cite{zhang2017beyond,guo2019toward} are specifically designed for image denoising.  Even though our method is designed for dealing with multiple degradations, it provides better performance than SOTA denoising methods.\\  

\noindent {\bf{Deblurring Experiments.}}  We compare the performance of NASFE against SOTA deblurring~\cite{PanECCV14,Nah_2017_CVPR,bulat2018super,ZiyiDeep,yasarla2020deblurring,kupyn2019deblurgan} methods using \textit{Test-B}. Methods such as \cite{PanECCV14,Nah_2017_CVPR,bulat2018super,ZiyiDeep,yasarla2020deblurring} use facial exemplar or semantic information as  priors to perform deblurring.  On the other hand, our network uses identity information as a prior to perform deblurring.   As can be seen from  Table~\ref{table:blur_comp}, our method outperforms the SOTA  method by 1.93dB in PSNR and 0.04 in SSIM.\\

\noindent {\bf{Low-light Enhancement Experiments.}}  We evaluated the performance of NASFE against SOTA low-light enhancement methods~\cite{fu2016weighted,guo2016lime,li2018structure,guo2020zero,wei2020physics} using \textit{Test-L}.   Even though our method is trained for addressing multiple degradations, NASFE  performed significantly better than \cite{fu2016weighted,guo2016lime,li2018structure,guo2020zero,wei2020physics}. As can be seen from Table~\ref{table:low_comp}, our method outperforms SOTA methods~\cite{fu2016weighted,guo2016lime,li2018structure,guo2020zero,wei2020physics}  by 3dB in PSNR and 0.10 in SSIM.  

More qualitative results are provided in the supplementary document.


\subsection{Multiple degradation experiments}
\noindent {\bf{Blur + Noise + Low-light Experiments.}}
We compare the performance of different methods on \textit{Test-BNL} which contains multiple degradations  blur, noise, and low-light conditions.  Note, we retrain \cite{bulat2018super,ZiyiDeep,yasarla2020deblurring,kupyn2019deblurgan,Li_2020_ECCV,Yang2020HiFaceGANFR} using degraded images that contain all degradations. Results are corresponding to this experiment are shown in Table~\ref{table:blur+noise+low_comp}.  As can be seen from this table, NASFE performs better by 2.1dB in PSNR and 0.07 in SSIM compared to second best performing method. Fig.~\ref{Fig:qual_BNL} shows the qualitative results of NASFE against other methods. The outputs of other methods are blurry or contain artifacts near eyes, nose and mouth.  On the other hand, NASFE produces clear and sharp face images. Furthermore, we can observe from Fig.~\ref{Fig:qual_BNL} and Fig.~\ref{Fig:qual_BNL_syn} that outputs of other methods still contain low-light conditions, whereas NASFE produces sharp face images with enhanced lighting conditions.

We conducted face recognition experiments using \textit{Test-BNL}, to show the significance of various face restoration methods on face recognition. Face recognition experiments are conducted using ArcFace~\cite{deng2019arcface} on \textit{Test-BN}, where Top-K similar faces for the restored face image are picked from the gallery set and used to compute accuracy. Table~\ref{table:blur+noise+low_comp} show the face recognition accuracies corresponding to different methods. We can clearly see that NASFE achieves an improvement of $7\%$ over the second best performing method.

\begin{table}[h!]
	\caption{PSNR$|$SSIM  comparisons for ablation study using \textit{Test-BN}, \textit{Test-BNL}. Learnable parameters are in millions.}
	\vskip-10pt
	\label{table:Ablation}
	\setlength{\tabcolsep}{8pt} 
	\renewcommand{\arraystretch}{1.25} 
	\resizebox{1.0\textwidth}{!}{
		\begin{tabular}{l|cc|cc|c}
			\hline
			\multirow{2}{*}{Method} & \multicolumn{2}{c|}{\textit{Test-BN}} & \multicolumn{2}{c|}{\textit{Test-BNL}} & \multirow{2}{*}{\begin{tabular}[c]{@{}c@{}}Learnable \\ Parameters\end{tabular}} \\ \cline{2-5}
			& CelebA & VGGFace2 & CelebA & VGGFace2 &  \\ \hline
			baseline network (BN) & 19.72$|$0.70 & 19.88$|$0.72 & 19.95$|$0.65 & 20.42$|$0.69 & 6.00 \\ \hline
			BN-NAS & 22.51$|$0.74 & 22.62$|$0.75 & 21.17$|$0.70 & 21.47$|$0.74 & 6.25 \\ \hline
			+ classification network & 23.28$|$0.76 & 23.04$|$0.76 & 21.56$|$0.72 & 21.95$|$0.76 & 6.25 \\ \hline
			+ identity informaation $I_{iden}$ & 24.88$|$0.81 & 24.60$|$0.82 & 23.06$|$0.80 & 23.47$|$0.80 & 6.40 \\ \hline
			NASFE (w/ $\mathcal{L}_{mse}$  and $\mathcal{L}_{per}$ ) & 25.10$|$0.84 & 24.92$|$0.84 & 23.35$|$0.83 & 23.76$|$0.83 & 6.40 \\ \hline
			NASFE w/ $\mathcal{L}_{final}$ & 25.57$|$0.87 & 25.49$|$0.87 & 23.80$|$0.85 & 24.15$|$0.86 & 6.40 \\ \hline
		\end{tabular}
	}
\end{table}

\subsection{Ablation study}
We conduct ablation studies using the test-sets \textit{Test-BN}, and \textit{Test-BNL} to show the improvements achieved by the different components in NASFE. We start with the baseline network (BN), and define it as a combination of three encoders ($E_B$, $E_N$, and $E_L$), a fusion network (composed of 4 Res2Blocks~\cite{gao2019res2net}), and a decoder ($De$). As shown in Table~\ref{table:Ablation}, BN performs very poorly due to its inability in processing task-specific features efficiently. Now, we introduce network architecture search in the fusion network by using fusion cells in-order to efficiently processing the task-specific features. The use of  network architecture search results in improvement of BN-NAS by $\sim 2$dB compared to BN. Then, we use class labels $c$ (computed using classification network) as input to BN-NAS which increases the performance of the network by $\sim 0.5$dB. Now we use the proposed identity information $I_{iden}$ of the identity present in the degraded image (refer to section~\ref{sec:iden}) which further improves the performance of the network by $\sim1.5$dB. The resultant network corresponds to NASFE. Note that BN and BN-NAS are trained using $\mathcal{L}_{mse}$. Now we train NASFE with $\mathcal{L}_{mse}$ and $\mathcal{L}_{per}$ which further improves the performance by 0.3dB. Now we use the proposed $\mathcal{L}_{iden}$ to construct $\mathcal{L}_{final}$ and train NASFE. The proposed $\mathcal{L}_{iden}$ improves the performance of NASFE by $\sim 0.5$dB.
\section{Conclusion}
We proposed a multi-task face restoration network, called NASFE, that can enhance poor quality face images containing a single degradation (i.e. noise or blur) or multiple degradations (noise+blur+low-light).  NASFE makes use of the clean face images of a person present in the degraded image to extract the identity information, and uses it to train the network weights. Additionally, we use network architecture search to design the fusion network in NASFE that  fuses the task-specific features obtained from the encoders. Extensive experiments shows that the proposed method performance significantly better than SOTA image restoration/enhancement methods on both synthetic degraded images as well as real-world images with multiple degradations (noise+blur+low-light).

{\small
\bibliographystyle{ieee_fullname}
\bibliography{egbib}
}

\end{document}